\documentclass[10pt]{article}

\usepackage{appendix}

\usepackage{fullpage}

\usepackage{amsmath,amssymb,amsfonts,amsthm}
\usepackage{optidef}
\usepackage{nicefrac}

\usepackage{float}
\floatstyle{ruled} 
\restylefloat{figure}

\usepackage[T1]{fontenc}
\usepackage{microtype}

\usepackage{graphicx}

\usepackage{booktabs}

\usepackage{siunitx}
\usepackage{tikz}
\usetikzlibrary{positioning}
\usepackage{pgfplots}
\usepgfplotslibrary{fillbetween}
\usepgfplotslibrary{groupplots}
\usepgfplotslibrary{colormaps}
\usepgfplotslibrary{colorbrewer}
\pgfplotsset{compat=1.3}
\pgfplotsset{
    cycle list/Set1-5,
    cycle multiindex* list={
        mark list*\nextlist
        Set1-5\nextlist
        linestyles\nextlist
    }
}

\usetikzlibrary{external}
\usepackage{natbib}
\setcitestyle{authoryear,round,citesep={;},aysep={,},yysep={;}}
\renewcommand{\cite}[1]{\citep{#1}}

\usepackage{doi} 

\usepackage{hyperref}
\hypersetup{
  colorlinks = true,
  allcolors = blue,
}
\usepackage[capitalise]{cleveref}
\newcommand{\letterref}[2]{\namecref{#1}~\hyperref[#1]{\ref*{#1}#2}}

\newcommand{\parheading}[1]{\vspace{2mm}\noindent\textit{#1}}

\newtheorem{prop}{Proposition}
\newtheorem{theorem}{Theorem}
\newtheorem{lemma}{Lemma}
\newtheorem{coro}{Corollary}

\newtheoremstyle{myremark}
{\topsep} 
{\topsep} 
{\normalfont} 
{} 
{\bfseries} 
{.} 
{5pt plus 1pt minus 1pt} 
{\thmname{#1}\thmnumber{ #2}\thmnote{ (#3)}} 

\theoremstyle{myremark}
\newtheorem{remark}{Remark}

\theoremstyle{definition}
\newtheorem{defn}{Definition}

\crefmultiformat{prop}{Propositions~#2#1#3}{ and~#2#1#3}{, #2#1#3}{ and~#2#1#3}
\crefmultiformat{theorem}{Theorems~#2#1#3}{ and~#2#1#3}{, #2#1#3}{ and~#2#1#3}
\crefmultiformat{lemma}{Lemmas~#2#1#3}{ and~#2#1#3}{, #2#1#3}{ and~#2#1#3}
\crefmultiformat{coro}{Corollaries~#2#1#3}{ and~#2#1#3}{, #2#1#3}{ and~#2#1#3}
\crefmultiformat{problem}{Problems~#2#1#3}{ and~#2#1#3}{, #2#1#3}{ and~#2#1#3}
\crefmultiformat{assump}{Assumptions~#2#1#3}{ and~#2#1#3}{, #2#1#3}{ and~#2#1#3}

\newcommand{\rank}{\mathrm{rank}}
\newcommand{\trace}{\mathrm{tr}}

\newcommand{\QNN}{\mathrm{QNN}}
\newcommand{\RUB}[3]{\mathrm{RUB}(#1,#2,#3)}
\newcommand{\SRUB}[3]{\mathrm{SRUB}(#1,#2,#3)}

\newcommand{\iid}{\textit{i.i.d.}}
\newcommand{\ie}{\textit{i.e.}}
\newcommand{\eg}{\textit{e.g.}}

\newcommand{\inv}{{-1}}
\newcommand{\invt}{{-\top}}

\newcommand{\bigO}{\mathcal{O}}

\newcommand{\R}{\mathbb{R}}
\newcommand{\Rn}[1]{\mathbb{R}^{#1}}
\newcommand{\Rnn}[1]{\mathbb{R}^{#1\times #1}}
\newcommand{\Rnp}[2]{\mathbb{R}^{#1\times #2}}
\newcommand{\Snn}[1]{\mathbb{S}^{#1\times #1}}

\newcommand{\bA}{\mathbf{A}}
\newcommand{\bB}{\mathbf{B}}
\newcommand{\bC}{\mathbf{C}}

\newcommand{\bI}{\mathbf{I}}

\newcommand{\bU}{\mathbf{U}}
\newcommand{\bV}{\mathbf{V}}

\newcommand{\bX}{\mathbf{X}}
\newcommand{\bY}{\mathbf{Y}}

\newcommand{\boldeta}{\boldsymbol{\eta}}

\newcommand{\bSigma}{\boldsymbol{\Sigma}}
\newcommand{\bLambda}{\boldsymbol{\Lambda}}
\newcommand{\bDelta}{\boldsymbol{\Delta}}

\newcommand{\cA}{\mathcal{A}}

\newcommand{\cP}{\mathcal{P}}

\newcommand{\cZ}{\mathcal{Z}}

\newcommand{\bb}{\mathbf{b}}

\newcommand{\bu}{\mathbf{u}}
\newcommand{\bv}{\mathbf{v}}
\newcommand{\bw}{\mathbf{w}}
\newcommand{\bx}{\mathbf{x}}
\newcommand{\by}{\mathbf{y}}
\newcommand{\bz}{\mathbf{z}}

\newcommand{\sigmax}{\sigma_{\mathrm{max}}}
\newcommand{\sigmin}{\sigma_{\mathrm{min}}}

\usepackage{color}

\title{Rank-One Measurements of Low-Rank PSD \\ Matrices Have Small Feasible Sets}
\author{
  T. Mitchell Roddenberry$^\dagger$,
  Santiago Segarra$^\dagger$,
  Anastasios Kyrillidis$^\ddag$ \vspace{1em}\\
  $^\dagger$Department of Electrical and Computer Engineering \\
  $^\ddag$Department of Computer Science \vspace{1em}\\
  Rice University, Houston, TX
}
\date{April 2021}

\begin{document}

\maketitle

\begin{remark}
  It was brought to our attention that a stronger version of~\cref{thm:recovery-guarantee} has been previously proven by~\citet{Kabanava:2016}.
  We suggest using said paper as a primary reference on this topic.
  We are leaving this manuscript online in case someone finds the alternative proof method interesting.
\end{remark}

\begin{abstract}
  We study the role of the constraint set in determining the solution to low-rank, positive semidefinite (PSD) matrix sensing problems.
  The setting we consider involves rank-one sensing matrices: 
  In particular, given a set of rank-one projections of an approximately low-rank PSD matrix, we characterize the radius of the set of PSD matrices that satisfy the measurements.
  This result yields a sampling rate to guarantee singleton solution sets when the true matrix is exactly low-rank, such that the choice of the objective function or the algorithm to be used is inconsequential in its recovery.
  We discuss applications of this contribution and compare it to recent literature regarding implicit regularization for similar problems.
  We demonstrate practical implications of this result by applying conic projection methods for PSD matrix recovery without incorporating low-rank regularization.
\end{abstract}

\section{Introduction}

We study the recovery of low-rank and approximately low-rank matrices from linear measurements.
In particular, for a linear map $\cZ:\Rnn{n}\to\Rn{m}$ and a positive semidefinite (PSD) matrix $\bX_0\in\Snn{n}$ where $\bb=\cZ(\bX_0)$, we characterize the optimization landscape in recovering $\bX_0$ from the measurements $\bb$.
When $m\ll n^2$, this problem is severely ill-posed and requires extra structure on the set of feasible solutions to successfully recover $\bX_0$.
Specifically, when $\bX_0$ is low-rank and $\cZ$ satisfies RIP~\citep{Chen:2015b} or RUB~\citep{Cai:2015} conditions, $\bX_0$ is the only low-rank solution to $\cZ(\bX)=\bb$.

In practice, finding the best low-rank solution to such a system is achieved via matrix factorization methods \cite{Tu:2016, zhao2015nonconvex, Zheng:2015, park2016provable, park2016finding, park2016non, bhojanapalli2016dropping, kyrillidis2018provable, ge2017no, hsieh2017non} or using convex nuclear norm penalties \cite{Recht:2010, candes2011robust}.
For PSD matrices, convex recovery problems and rank-constrained instances often take the form
\begin{argmini}
  {\small \bX\succeq 0}{f(\bX)}
  {\label{eq:general-recovery}}{\bX^* \in }
  \addConstraint{\cZ(\bX)}{=\bb,}
\end{argmini}
for some regularization function $f$, such as the nuclear norm of $\bX$ or the non-convex $\rank(\cdot)$ function.

Recovery of low-rank matrices has a wide array of applications, including recommendation systems \cite{Recht:2010, Candes:2009, davenport2016overview, chandrasekaran2012convex, chen2018harnessing, candes2011tight}, quantum state tomography \cite{Recht:2010, kyrillidis2018provable, flammia2012quantum, gross2010quantum, liu2011universal, Chen:2015b, Cai:2015},  phase retrieval and blind deconvolution \cite{shechtman2015phase, fienup1982phase, Candes:2013, Chen:2015b, Cai:2015, li2016low, segarra2017blind, ling2015self}, neural word embeddings \cite{mikolov2013distributed, pennington2014glove}, text classification \cite{joulin2017bag}, convexified convolutional NNs \cite{zhang2017convexified}, and SDP instances \cite{burer2003nonlinear, bhojanapalli2018smoothed, bhojanapalli2016dropping, kyrillidis2018provable, wang2017mixing, yurtsever2019scalable, goto2019combinatorial}.
Because of its importance in practice, there has been a large push in the literature to develop efficient algorithms for this task \cite{davenport2016overview}.

\subsection{Contribution}

In this work, we derive bounds on the size of the feasibility set when recovering low-rank and approximately low-rank PSD matrices from (potentially noisy) rank-one projections.
In particular, we show that the PSD constraint of the feasible set for many optimization problems is restrictive enough to render the chosen penalty function inconsequential for recovery, especially when the spectrum of the matrix decays quickly.
Importantly, this paper \emph{does not} propose a new algorithm or method: rather, it seeks to help explain the success of many matrix recovery methods by characterizing the optimization landscape on which they operate.

Although the focus of this work is theoretical in nature, it is relevant to the use of fast algorithms for finding solutions to low-rank matrix recovery problems.
In essence, if the objective function is not consequential, one can choose an objective function based on algorithmic properties, rather than based on the structure it imparts upon the optimal solution.

\subsection{Notation}

We refer to matrices using bold uppercase letters, \eg{} $\bA,\bB,\bC$, and to (column) vectors with bold lowercase letters, \eg{} $\bv,\bw,\bx$.
Entries of a matrix $\bA$ are indicated by $A_{ij}$, while those of vector $\bx$ are denoted by $x_i$.
For clarity, we alternatively use the notation $[\bx]_i=x_i$.
The $\ell_1$-norm of a vector is denoted by $\|\cdot\|_1$.
The nuclear and Frobenius norms of a matrix are denoted by $\|\cdot\|_*$ and $\|\cdot\|_F$, respectively.
The inverse of a matrix is indicated by $\bX^\inv$, and the matrix transpose is $\bX^\top$.
The inverse of the transposed matrix, or the transpose of the inverted matrix, is $\bX^\invt$.

For a matrix $\bX\in\Snn{n}$ with singular value decomposition $\bX=\sum_{i=1}^n\sigma_i\bu_i\bv_i^\top$, the minimum and maximum singular values of $\bX$ are denoted by $\sigmin(\bX)$ and $\sigmax(\bX)$, respectively.
The matrix constructed from the lower $p$ singular value/vector pairs is denoted by $[\bX]_{-p}$.
In particular, $\|[\bX]_{-p}\|_*$ is equal to the sum of the $p$ smallest singular values of $\bX$.

\subsection{Related Work}

\parheading{Low-Rank Regularization.} Matrix factorization for recovery of PSD matrices optimizes an objective function over a matrix factor $\bU\in\Rnp{n}{r}$, \ie{}
\begin{argmini}
  {\bU\in\Rnp{n}{r}}{f(\bU\bU^\top)}
  {\label{eq:matrix-factorization}}{\bU^* \in }
  \addConstraint{\cZ(\bU\bU^\top)}{=\bb.}
\end{argmini}
The choice of dimension $r$ constrains the recovered matrix $\bX^*=\bU^*\bU^{*\top}$ to be rank-$r$.
In particular, \citet{Zheng:2015, Tu:2016} study gradient descent algorithms when the constraint in \eqref{eq:matrix-factorization} is relaxed and included in $f$ as a least-squares loss on $\bU\bU^\top$.
\cite{park2016finding, bhojanapalli2016dropping} considers the more general case where $f$ is any convex function.

In the overparameterized regime where $r=n$, low-rankness of the final solution is often obtained via a nuclear norm penalty on the matrix $\bU\bU^\top$: equivalently, the squared Frobenius norm of $\bU$.
This has been applied with great success to the tasks of phase retrieval~\citep{Candes:2013, Candes:2015}, matrix sensing~\citep{Recht:2010}, and matrix completion~\citep{Candes:2009}.
Most related to our work, \citet{Kueng:2017} derived sampling requirements for the setting where $\cZ$ consists of random rank-one projections.

\parheading{Semidefinite Programming.} Optimization problems with PSD+affine constraints include semidefinite programs (SDP).
\citet{Pataki:2000, barvinok1995problems} thoroughly characterized the geometry of SDPs, including the cases where the feasible set is a singleton (as we do), albeit without statistical guarantees or sampling rates for random measurements.
\citet{Boumal:2020} also studied the landscape of factorized methods for finding low-rank solutions to SDPs, showing that for a sufficiently expressive Burer-Monteiro factorization $\bX=\bU\bU^\top$, global minima for the factor $\bU$ correspond to global minima in the matrix $\bX$; see also \cite{boumal2016non, pumir2018smoothed, bhojanapalli2018smoothed}.
In contrast, this work does not depend on a particular factorization, holding for \emph{any} formulation of an optimization problem with PSD+affine constraints, or even cases where PSDness is enforced implicitly via the Burer-Monteiro factorization, as long as there are sufficiently many random affine measurements.

\parheading{Algorithmic Regularization.} Recent works have shown that, in some cases, nuclear-norm penalties on the recovered matrix are not necessary when certain algorithms are used.
In particular, \citet{Gunasekar:2017} and \citet{Li:2018} studied the conditions under which gradient descent on matrix factors for an affine least-squares loss converges to the minimum nuclear norm solution, even in the overparameterized case.
These works characterize the optimization landscape of nonconvex objectives on the matrix factors, and show that carefully initialized descent methods remain in a well-behaved basin of low-rank solutions.
Unlike these approaches, our results are independent of the algorithm used to solve the optimization problem; see also \cite{razin2020implicit}.

\parheading{Problem-Statement Regularization.} Another direction in characterizing the landscape of matrix sensing problems is to bound the size of the constraint set.
In particular, if there is only one feasible solution to a set of constraints, then \emph{any} regularization scheme or algorithm---that operates within the constraint region---will find that point, no matter how nonconvex or overparameterized the problem setting is.
The only regularization needed, then, is that of the problem statement itself.

This was considered by \citet{Wang:2011}, where they show that random sensing matrices drawn from a Gaussian ensemble yield uniquely feasible PSD recovery programs when $m\in\bigO(n^2)$.
Another step in this direction was made by \citet{Demanet:2014}, where they derive $\bigO(n\log n)$ sampling rates for unique recovery of rank-one PSD matrices in the phase retrieval setting.
This rate was tightened to $\bigO(n)$ by \citet{Candes:2014}.
\citet{Geyer:2020} showed a similar result for general low-rank PSD matrices under Wishart sensing maps, deriving $\bigO(nr)$ sampling rates for unique PSD matrix recovery.
More broadly, the idea of bounding the size of the feasible set is analogous to work in nonnegative compressive sensing.
\citet{Bruckstein:2008} showed conditions under which compressive sensing of sparse, nonnegative signals requires no regularization.

\section{Rank-One Matrix Sensing}

We consider measurements of a PSD matrix taken via \emph{rank-one projections}.
That is, consider a symmetric matrix $\bX_0\in\Snn{n}$ such that $\bX_0\succeq 0$, and a \emph{sensing map} $\cZ:\Snn{n}\to\Rn{m}$.
The sensing map $\cZ$ is defined by a set of $m$ \emph{sensing vectors} $\{\bz_i\}_{i=1}^m$, so that
\begin{equation}\label{eq:sensing-map}
  b_i=\cZ_i(\bX)=\bz_i^\top\bX\bz_i.
\end{equation}
In particular, let $\bb=\cZ(\bX_0)$ be the vector of measurements obtained by applying $\cZ$ to the matrix of interest, $\bX_0$.
Or, of more practical interest, let us consider measurements in the presence of noise, \ie{},
\begin{equation}\label{eq:noisy-sensing-map}
  b_i=\cZ_i(\bX)+\eta_i,
\end{equation}
so that $\bb=\cZ(\bX_0)+\boldeta$.

The problem, then, is to recover the matrix $\bX_0$ given the measurement vector $\bb$ and the sensing vectors $\{\bz_i\}_{i=1}^m$.
This is then solved via the program~\eqref{eq:general-recovery}, with an appropriate relaxation of the affine constraint to account for the noise $\boldeta$.
A common assumption is that $\bX_0$ is low-rank, yielding the nuclear norm convex relaxation $f(\bX)=\|\bX\|_*$.

Many papers have considered the conditions under which $\bX_0$ can be recovered, particularly when using the nuclear norm penalty~\citep{Chen:2015b,Cai:2015,Candes:2009,Recht:2010}.
By establishing RIP-type conditions on the sensing map $\cZ$, nuclear norm minimization can be shown to recover $\bX_0$ exactly for sufficiently large $m$, under assumptions on the distribution of the sensing vectors $\{\bz_i\}_{i=1}^m$.

In this work, we establish conditions under which nuclear norm minimization is unnecessary, in the spirit of \citet{Demanet:2014} and \citet{Geyer:2020}.
That is, if we take the criterion $f$ to be constant-valued, thereby reducing~\eqref{eq:general-recovery} to a feasibility problem, under what sampling conditions can we expect ``good'' recovery of the matrix $\bX_0$?
This is distinct from the work on implicit regularization for these problems, such as that of \citet{Gunasekar:2017} and \citet{Li:2018}, which shows that gradient descent for nonconvex matrix factorization favors low-rank solutions.
Instead, we aim to characterize the behavior of \emph{any algorithm} that finds feasible PSD matrices, no matter what regularization is used.

\subsection{A Numerical Surprise}

\begin{figure}[tb]
    \centering
    \includegraphics[width=0.7\linewidth]{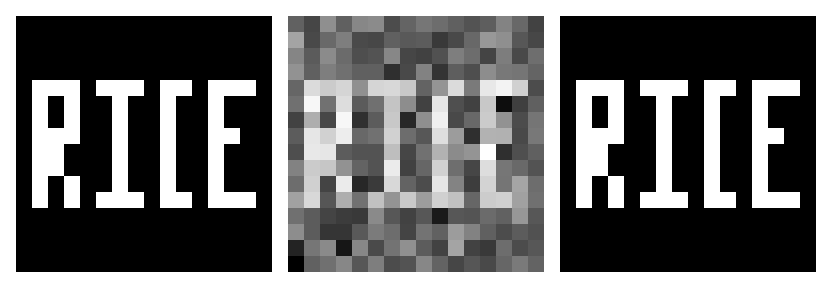}
    \caption{
    Success of PSD constraints in phase retrieval.
    \textbf{(Left)} Original vector $\bx_0\in\Rn{256}$.
    \textbf{(Center)} Top eigenvector of $\bX$ recovered according to~\eqref{eq:demo-nopsd}. 
    \textbf{(Right)} Top eigenvector of $\bX$ recovered according to~\eqref{eq:demo-psd}.
    }
    \label{fig:demo}
\end{figure}

We begin by considering a simple example in real-valued phase retrieval.
Let $\bx_0\in\Rn{n}$ be a signal from which we take random measurements of the form
\begin{equation}\label{eq:demo-phase-retrieval}
  b_i=|\langle\bx_0,\bz_i\rangle|^2,\ i=1,\ldots,m,
\end{equation}
for random vectors $\{\bz_i\}_{i=1}^m$.
To recover $\bx_0$ from $\bb\in\Rn{m}$, we lift~\eqref{eq:demo-phase-retrieval} and find a matrix $\bX$ such that $\bz_i^\top\bX\bz_i=b_i$ for all $i=1,\ldots,m$.
This lifting stems from the fact that $\bX_0=\bx_0\bx_0^\top$ satisfies this condition, so we estimate $\bx_0$ by the top eigenvector of the recovered matrix.
This is the well-known \emph{phase retrieval problem}, and is often solved using this lifting approach coupled with a convex penalty, such as the trace of $\bX$~\citep{Candes:2013}.

We consider the case where no such penalties are used.
In~\letterref{fig:demo}{ (Left)}, we see a simple vector $\bx_0\in\Rn{256}$, representing a $16\times 16$ grayscale image.
We then take $m=2560$ measurements of the form~\eqref{eq:demo-phase-retrieval}, and solve two variants of the lifted feasibility problem.

In~\letterref{fig:demo}{ (Center)}, we solve the feasibility program
\begin{equation}\label{eq:demo-nopsd}
    \begin{aligned}
      &\mathrm{find}\quad\bX \quad
      &\mathrm{subject~to}\quad\cZ(\bX)=\bb.
    \end{aligned}
\end{equation}
Observe that the solution to~\eqref{eq:demo-nopsd} is not unique in this underdetermined sampling case, since $m\ll n^2$.
We project the all-zeros matrix onto the feasible set, implicitly finding a solution with low Frobenius norm, and display the top eigenvector of the recovered matrix.
Although there is a semblance of the original image here, it is corrupted by noise lying in the null space of $\cZ$.
Typically, this would be alleviated by incorporating the knowledge that $\bX_0$ is rank-one, motivating the use of nuclear norm penalties to find low-rank solutions.

Another form of regularization can come from the fact that $\bX_0$ is PSD.
In~\letterref{fig:demo}{ (Right)}, we solve the same program with a restricted domain:
\begin{equation}\label{eq:demo-psd}
    \begin{aligned}
        &\mathrm{find}\quad\bX\succeq 0 \quad
        &\mathrm{subject~to}\quad\cZ(\bX)=\bb.
    \end{aligned}
\end{equation}
At first glance, this may also appear ill-posed, but we surprisingly attain near-perfect recovery of the original image.
This suggests that under sufficient sampling conditions on a low-rank matrix, \emph{enforcing PSDness shrinks the feasible set to a singleton.}
Indeed, \citet[Theorem~1.3]{Candes:2014} shows that when $m\geq Cn$ for some sufficiently large $C$, the phase retrieval problem has a unique feasible solution.
Next, we present our main theorem, which subsumes this result as a special case.

\section{Main Result}

For (potentially noisy) matrix sensing of a PSD matrix with random rank-one projections, we state the following sampling result for matrix recovery.
\begin{theorem}\label{thm:recovery-guarantee}
  For a collection of $m$ sensing vectors $\{\bz_i\}_{i=1}^m$ drawn \iid{} uniformly over the unit sphere, let the corresponding sensing map be $\cZ:\Snn{n}\to\Rn{m}$.
  For all PSD matrices $\bX_0$, with probability at least $1-\bigO(\exp(-\gamma n))$, if $m>Crn$, any $\bX\succeq 0$ such that $\|\cZ(\bX)-\bb\|_1\leq\|\boldeta\|_1$, with $\bb,\boldeta$ as in \eqref{eq:noisy-sensing-map}, will satisfy
  \begin{equation}\label{eq:recovery-guarantee}
    \|\bX-\bX_0\|_F\leq c_1\frac{\|[\bX_0]_{-(n-r)}\|_*}{\sqrt{r}} + c_2\frac{\|\boldeta\|_1}{m},
  \end{equation}
  for constants $C,c_1,c_2,\gamma$.
\end{theorem}

The proof is deferred to~\cref{S:proof_main_theorem}. \Cref{thm:recovery-guarantee} bounds the radius of the set of PSD matrices $\bX$ such that $\cZ(\bX)\approx\cZ(\bX_0)$ in terms of the rate of decay of the lower eigenvalues of $\bX_0$ and the measurement noise.
Indeed, since~\cref{thm:recovery-guarantee} is stated with full-rank $\bX_0$ in mind, if $\rank(\bX_0)=r$, noting that $\|[\bX_0]_{-(n-r)}\|_*=0$ immediately yields the following:
\begin{coro}\label{cor:lowrank-recovery-guarantee}
  Under the same conditions as~\cref{thm:recovery-guarantee}, if the matrix $\rank(\bX_0)=r$, then the set of PSD matrices such that $\|\cZ(\bX)-\bb\|_1\leq\|\boldeta\|_1$ has radius in the Frobenius norm bounded by $c_2\|\boldeta\|_1/m$, for some constant $c_2$.
\end{coro}

\Cref{cor:lowrank-recovery-guarantee} demands an $\bigO(nr)$ sampling rate for exact reconstruction of rank-$r$ PSD matrices in the noiseless case, when no regularization is used (\eg{} nuclear norm minimization).
This is an asymptotically tight result, as there are $nr$ degrees of freedom for low-rank PSD matrices, exemplified by the factorization $\bX_0=\bU\bU^\top$, where $\bU\in\Rnp{n}{r}$.
Interestingly, this matches the state-of-the-art for recovery of general Hermitian matrices from rank-one measurements using nuclear norm penalties and no semidefinite constraints, as in \citet[Theorem 2]{Kueng:2017} which requires $\bigO(nr)$ samples for low-rank recovery with nuclear norm minimization.

\begin{remark}
\Cref{thm:recovery-guarantee} is stated for sensing vectors distributed on the unit sphere, but this is largely for convenience in the proof.
The setting where $\bz\sim\mathcal{N}(0,\bI)$ is essentially the same, since the measurements can be scaled to obtain a similar program, as noted by \citet{Candes:2013,Candes:2014}.
\end{remark}

\section{Applications}

Before moving on to the proof of \cref{thm:recovery-guarantee}, we discuss a few applications of our result within the context of existing literature on implicit and explicit regularization methods.
A common theme is that the specialization of our main result to these domains recovers and strengthens state-of-the-art application-specific results.

\subsection{No Algorithmic Regularization Needed when Training Quadratic Neural Networks}

We consider the training of single-layer neural networks with quadratic activations (QNNs).
Such a neural network is parameterized by a matrix $\bU\in\Rnp{n}{p}$, with real-valued output obtained on input vectors $\bx\in\Rn{n}$ according to
\begin{equation}\label{eq:qnn-def}
  \QNN_{\bU}(\bx) = {\mathbf 1}^\top q(\bU^\top\bx),
\end{equation}
where $q(\cdot)$ squares a vector elementwise, and ${\mathbf 1}\in\Rn{p}$ is the all-ones vector.
Observe that evaluating a QNN parameterized by $\bU$ is equivalent to a rank-one measurement of the PSD matrix $\bU\bU^\top$:
\begin{equation}\label{eq:qnn-is-matrix-sensing}
  \QNN_{\bU}(\bx) = \bx^\top(\bU\bU^\top)\bx.
\end{equation}

For a dataset $\{\bx_i,y_i\}_{i=1}^m$ that can be fit perfectly by a QNN, there is a matrix $\bU_0$ such that $\QNN_{\bU_0}(\bx_i)=y_i$ for all $1\leq i\leq m$.
Then, the QNN could be viewed as a generating process for the dataset, from which we recover the parameters $\bU_0$ by sampling.
We apply~\cref{cor:lowrank-recovery-guarantee} to characterize the sample complexity for recovery of $\bU_0$ (up to a unitary rotation) using \emph{any} algorithm that can exactly fit the given data.
\begin{prop}\label{prop:qnn-lowrank-uniqueness}
  For a dataset $\{\bx_i,y_i\}_{i=1}^m$ where the vectors $\bx_i\in\Rn{n}$ are drawn \iid{} from a uniform spherical distribution, and rank-$r$ matrix $\bU_0$ such that $\QNN_{\bU_0}(\bx_i)=y_i$ for all $1\leq i\leq m$, let $\bU\in\Rnp{n}{p}$ with $p\geq r$ be the parameter matrix yielded by any algorithm where $\QNN_{\bU}(\bx_i)=y_i$ for all $1\leq i\leq m$.
  Then, there exists a constant $C$ such that if $m>Cnr$, then $\bU\bU^\top=\bU_0\bU_0^\top$ with high probability.
\end{prop}

In the simple case where $\bU_0$ is exactly low-rank, \cref{prop:qnn-lowrank-uniqueness} guarantees \emph{exact recovery} of $\bU_0$ (up to a unitary rotation) when the number of samples $m\in\bigO(nr)$, even when $\bU$ is allowed to be full-rank (\ie{} the overparameterized regime).
This is in contrast to the results of~\citet{Li:2018}, where $m\in\widetilde{\bigO}(nr^2)$ samples are required for low-rank recovery in the overparameterized setting, as well as a particular gradient descent algorithm that is sensitive to initialization.
Our characterization of this problem setting in~\cref{prop:qnn-lowrank-uniqueness} requires fewer samples asymptotically, as well as being agnostic to the particular algorithm used in recovering the matrix $\bU$.

\subsection{Phase Retrieval Only Needs $O(n)$ Measurements}

\Cref{thm:recovery-guarantee} naturally applies to the (real-valued) \emph{phase retrieval problem}.
That is, we aim to recover a vector $\bx_0\in\Rn{n}$ from a set of measurements
\begin{equation}\label{eq:phase-retrieval-map}
  b_i=|\langle\bx_0,\bz_i\rangle|^2+\eta_i,\ i=1,\ldots,m,
\end{equation}
again for $\bz_i$ drawn \iid{} uniformly on the unit sphere.

Results similar to ours were shown by~\citet{Candes:2014}, where they proved that lifting the phase retrieval problem to a PSD matrix recovery problem yields, with high probability, a singleton (or small) feasibility set, as long as $m\geq Cn$.
That is, for $\cZ(\bx_0\bx_0^\top)=\bb$, they make the following guarantee:
\begin{theorem}[{\citet[Theorem~1.3]{Candes:2014}}]\label{thm:phase-retrieval}
  Let $m\geq Cn$ for some sufficiently large $C$.
  Then, if $\bX\succeq 0$ minimizes $\|\cZ(\bX)-\bb\|_1$ with $\bb$ as in \eqref{eq:phase-retrieval-map}, we have with probability at least $1-\bigO(\exp({-\gamma m}))$
  \begin{equation}
      \|\bX-\bx_0\bx_0^\top\|_F\leq C_0\frac{\|\boldeta\|_1}{m},
  \end{equation}
  for constants $C_0,\gamma$.
\end{theorem}
Specializing our result to $\bX_0=\bx_0\bx_0^\top$ (\ie{}, $r=1$) yields the same linear sampling rate via \cref{cor:lowrank-recovery-guarantee}.
Moreover, \cref{cor:lowrank-recovery-guarantee} is stated for \emph{any} feasible matrix, not just the one closest to the set of measurements in the $\ell_1$-norm, as specified in~\cref{thm:phase-retrieval}.

\subsection{Covariance Estimation from Quadratic Sampling}

Another application of rank-one matrix sensing is in covariance estimation: this can be viewed as a higher-rank version of the phase retrieval problem.
In particular, \citet[Theorem~1]{Chen:2015b} derive a similar rate to ours using a trace minimization program.
\begin{theorem}[{\citet[Theorem~1]{Chen:2015b}}]\label{thm:covariance-trace-minimization}
    For the measurement model in \eqref{eq:noisy-sensing-map}, the minimum nuclear-norm matrix $\bX\succeq 0$ such that $\|\cZ(\bX)-\bb\|_1\leq\|\boldeta\|_1$ satisfies, with probability at least $1-\bigO(\exp(-\gamma m))$,
    \begin{equation}
        \|\bX-\bX_0\|_F\leq c_1\frac{\|[\bX_0]_{-(n-r)}\|_*}{\sqrt{r}} + c_2\frac{\|\boldeta\|_1}{m},
    \end{equation}
    simultaneously for all $\bX_0\succeq 0$, provided that $m>Cnr$, where $C,c_1,c_2,\gamma$ are positive constants.
\end{theorem}
This error rate is the same as~\cref{thm:recovery-guarantee}, indicating that trace minimization only improves sampling rates by a constant factor.

\section{Proof of~\Cref{thm:recovery-guarantee}}
\label{S:proof_main_theorem}

Our proof of~\cref{thm:recovery-guarantee} will proceed as follows.
We first form a bijection between the feasibility set of~\eqref{eq:general-recovery} and another subset of the PSD cone, as well as an associated sensing map that preserves the measurement vector.
We then show that the elements of this transformed set have approximately equal nuclear norm.
Additionally, we apply guarantees of an RIP-type condition for the transformed sensing map, thereby bounding the radius of the transformed set, which finally bounds the radius of the original feasible set.

\subsection{Establishing Well-Behaved Coordinates}

Define $\bSigma=\frac{1}{m}\sum_{i=1}^m\bz_i\bz_i^\top$ so that $\bSigma\succ 0$ when $m\geq n$, with probability 1%
\footnote{Any weighting of the terms here is suitable, \ie{} $\bSigma=\sum_{i=1}^m\phi_i\bz_i\bz_i^\top$ as done by \citet{Geyer:2020}, as long as $\{\phi_i\}_{i=1}^m$ is such that $\bSigma\succ 0$.}.
Thus, $\bSigma$ admits the decomposition $\bSigma=\bV\bV^\top$, where $\bV$ has eigenvalues equal to the square root of the eigenvalues of $\bSigma$.
Therefore, $\bV^\inv$ exists, from which we define a transformed sensing map $\widehat{\cZ}:\Snn{n}\to\R^m$ as
\begin{equation}\label{eq:transformed-sensing-map}
  \widehat{\cZ}_i(\bY)=(\bV^\inv\bz_i)^\top\bY(\bV^\inv\bz_i).
\end{equation}
Relating $\cZ$ and $\widehat{\cZ}$ is a natural bijection $g$ from the PSD cone to itself, \ie{},
\begin{equation}\label{eq:psd-bijection}
  \begin{aligned}
    g(\bX)&=\bV^\top\bX\bV \\
    g^{-1}(\bY)&=\bV^\invt\bY\bV^\inv.
  \end{aligned}
\end{equation}
From the definition of $g$ and $\widehat{\cZ}$, we have for all $\bX\in\Snn{n}$ that $\cZ(\bX)=\widehat{\cZ}(g(\bX))$, so that for each $\bX$, there is a corresponding $\bY=g(\bX)$ such that $\widehat{\cZ}(\bY)=\cZ(\bX)$, with the converse also holding due to $g$ being a bijection.

We now consider the feasible sets for both $\cZ$ and $\widehat{\cZ}$, defined respectively as
\begin{equation}\label{eq:feasibility-sets}
  \begin{aligned}
    F&=\{\bX\succeq 0:\|\cZ(\bX)-\bb\|_1\leq\|\boldeta\|_1\}, \\
    \widehat{F}&=\{\bY\succeq 0:\|\widehat{\cZ}(\bY)-\bb\|_1\leq\|\boldeta\|_1\},
  \end{aligned}
\end{equation}
where $g$ forms a bijection between $F$ and $\widehat{F}$.
In particular, $\bX_0\in F$ and $g(\bX_0)\in\widehat{F}$, by construction.
It can be shown that the nuclear norm is approximately flat in the transformed feasible set:
\begin{lemma}\label{lem:fixed-trace}
  All matrices $\bY\in\widehat{F}$ satisfy
  \begin{equation}
      \frac{\sum_{i=1}^m b_i-\|\boldeta\|_1}{m}\leq\|\bY\|_*\leq\frac{\sum_{i=1}^m b_i+\|\boldeta\|_1}{m} 
  \end{equation}
\end{lemma}
The proof of this result mirrors that of~\citet{Geyer:2020}, so we relegate it to \cref{app:flat-nuc}.

\subsection{Recovery Guarantees in $\widehat{F}$}

The transformation $g:F\to\widehat{F}$ yields a coordinate system in which the matrix recovery problem is amenable to analysis.
To this end, we introduce the notion of a sensing map fulfilling the \emph{Symmetrized Restricted Uniform Boundedness property}:
\begin{defn}\label{defn:srub}
    A symmetric sensing map $\cZ:\Snn{n}\to\Rn{m}$ is said to fulfill the \emph{Symmetrized Restricted Uniform Boundedness property} (SRUB) of order $r$ with constants $C_1,C_2$ if for all symmetric rank-$r$ matrices $\bX\in\Snn{n}$, the map $\cZ'_i(\bX)=\frac{1}{2}(\cZ_{2i-1}(\bX)-\cZ_{2i}(\bX)), i=1,\ldots,\lfloor\frac{m}{2}\rfloor$ satisfies
    \begin{equation}\label{eq:srub-ugly}
        C_1\leq
        \frac{\|\cZ'(\bX)\|_1/\lfloor\frac{m}{2}\rfloor}
        {\|\bX\|_F}
        \leq C_2.
    \end{equation}
    We denote the set of all such symmetric sensing maps by $\SRUB{r}{C_1}{C_2}$.
\end{defn}
\Cref{defn:srub} is based on an intermediate construction used by \citet{Cai:2015}, and is discussed in more depth in \cref{app:srop,app:srub}.
In particular, it allows us to inherit recovery guarantees granted by sensing maps that obey the Restricted Uniform Boundedness (RUB) property, as in \citet[Section~2]{Cai:2015}.

To characterize the landscape of $\widehat{F}$, we develop appropriate sampling conditions for $\widehat{\cZ}$ to fulfill the SRUB property with appropriate constants.
\begin{prop}\label{prop:basic-rub-sampling}
  Set parameters $k\geq 2$, $C_1<1/3$, and $C_2>1$ arbitrarily.
  Then, for constants $C$ and $\delta$ dependent on these three parameters, if $m>Cnr$ then $\cZ\in\SRUB{2kr}{C_1}{C_2}$ with probability at least $1-\exp(-m\delta)$.
\end{prop}
\Cref{prop:basic-rub-sampling} is stated in the proof of \citet[Proposition 3.1]{Cai:2015}, so we omit the proof here.

\begin{prop}\label{prop:basic-sing-sampling}
  For some constants $C,\gamma$, if $m>Cn$, then
  \begin{equation}
    2\sqrt{2}-2\leq\sigmin(\bSigma)\leq\sigmax(\bSigma)\leq 4-2\sqrt{2}
  \end{equation}
  with probability at least $1-\exp(-n\gamma)$.
  In particular,
  \begin{equation}
    \frac{\sigmax(\bSigma)}{\sigmin(\bSigma)}\leq\sqrt{2}.
  \end{equation}
\end{prop}
\Cref{prop:basic-sing-sampling} follows from~\citet[Corollary~5.52]{Vershynin:2010}, derived in \cref{app:sing-proof}.
These results allow us to control the properties of $\widehat{\cZ}$, in order to yield strong recovery guarantees in $\widehat{F}$.
We begin by leveraging \cref{prop:basic-rub-sampling,prop:basic-sing-sampling} to establish the SRUB property of $\widehat{\cZ}$ and the spectral properties of $\bSigma$.
\begin{lemma}\label{lem:transformed-rub-guarantee}
  If for some constant $C$, it holds that $m>Cnr$, then the following holds with probability at least $1-\bigO(\exp(-n\gamma))$:
  \begin{gather}
    \widehat{\cZ}\in\SRUB{2kr}{C_1}{C_2}, \quad C_2/C_1<\sqrt{2k} \\
    \sigmin(\bSigma)\geq 2\sqrt{2}-2 \\
    \frac{\sigmax(\bSigma)}{\sigmin(\bSigma)}\leq\sqrt{2}.
  \end{gather}
\end{lemma}
The proof of \cref{lem:transformed-rub-guarantee} is deferred to \cref{app:trans-srub}.
With \cref{lem:transformed-rub-guarantee}, we can make recovery guarantees in $\widehat{F}$ under the sensing map $\widehat{\cZ}$.
To see this, take any element $\bY_1\in\widehat{F}$.
We then have the following guarantee on the quality of $\bY_1$ with respect to $\bY_0=g(\bX_0)$.
Under the conditions of \cref{lem:transformed-rub-guarantee}, by \citet[Lemma~7.8]{Cai:2015}, for some constants $c_1,c_2$,
\begin{equation}\label{eq:transformed-rub-guarantee}
  \begin{gathered}
  \|\bY_1-\bY_0\|_F\leq c_1\frac{\xi}{\sqrt{r}}+c_2\frac{\|\boldeta\|_1}{m}, \\
  \xi := \max\left(\|[\bY_0]_{-(n-r)}\|_*, \|[\bY_1]_{-(n-r)}\|_*\right).
  \end{gathered}
\end{equation}

\subsection{Distortion Induced by $g$}

We bound the radius of $F$ about $\bX_0$ in the Frobenius norm by considering the radius of $\widehat{F}$ about $\bY_0=g(\bX_0)$.
Let $\bY$ be an arbitrary element of $\widehat{F}$, so there exists an $\bX\in F$ such that $\bX=g^\inv(\bY)=\bV^\invt\bY\bV^\inv$.
Define $\bDelta_\bY=\bY-\bY_0$.
It follows that
\begin{equation}\label{eq:error-distortion}
  \|\bX-\bX_0\|_F=\|\bV^\invt\bDelta_\bY\bV^\inv\|_F\leq\frac{\|\bDelta_\bY\|_F}{\sigmin(\bSigma)}.
\end{equation}
An immediate implication of~\eqref{eq:error-distortion} is:
\begin{lemma}\label{lem:radius-distortion}
  The radius of $F$ about $\bX_0$ in the Frobenius norm is bounded by the radius of $\widehat{F}$ about $\bY_0=g(\bX_0)$ divided by the minimum eigenvalue of $\bSigma$.
\end{lemma}

\subsection{Concluding the Proof}

We now establish \cref{thm:recovery-guarantee}.
Let $C$ be a constant such that $m>Cnr$ satisfies the sampling condition of~\cref{lem:transformed-rub-guarantee}.
Then, the difference in the Frobenius norm between any $\bY\in\widehat{F}$ and $\bY_0=g(\bX_0)$ is bounded according to~\eqref{eq:transformed-rub-guarantee}.
\Cref{lem:fixed-trace} establishes that every element of $\widehat{F}$ has nuclear norm contained in a small interval, thereby providing a bound on the radius of $\widehat{F}$ about $\bY_0$.
In particular, \cref{lem:fixed-trace} implies that $\xi$ in~\eqref{eq:transformed-rub-guarantee} is upper bounded by $\|[\bY_0]_{-(n-r)}\|_*+2\|\boldeta\|_1/m$.
Therefore, for all $\bY\in\widehat{F}$,
\begin{align}
    \|\bY-\bY_0\|_F &\leq c_1\frac{\|[\bY_0]_{-(n-r)}\|_*+2\|\boldeta\|_1/m}{\sqrt{r}} +c_2\frac{\|\boldeta\|_1}{m} \nonumber \\
    &\leq c_1\frac{\|[\bY_0]_{-(n-r)}\|_*}{\sqrt{r}} \nonumber \\
    &\quad+\left(2c_1+c_2\right)\frac{\|\boldeta\|_1}{m}. \label{eq:transformed-universal}
\end{align}

Recalling that $\bY_0=g(\bX_0)$ by definition, we have
\begin{equation}\label{eq:lower-nuclear-bound}
    \|[\bY_0]_{-(n-r)}\|_*  \leq \sigmax(\bSigma)\|[\bX_0]_{-(n-r)}\|_*,
\end{equation}
following from the result of \citet[Theorem 2]{Wang:1997}, derived in \cref{app:lower-nuc}.

Finally, applying~\cref{lem:radius-distortion} yields, for all $\bX\in F$,
\begin{align}
    \|\bX-\bX_0\|_F &\leq c_1
    \frac{\sigmax(\bSigma)}{\sigmin(\bSigma)}
    \frac{\|[\bX_0]_{-(n-r)}\|_*}{\sqrt{r}} \nonumber \\
                    &\quad +\left(\frac{2c_1+c_2}{\sigmin(\bSigma)}\right)\frac{\|\boldeta\|_1}{m} \nonumber \\
                    &\overset{(a)}{\leq} c_1\sqrt{2}\frac{\|[\bX_0]_{-(n-r)}\|_*}{\sqrt{r}} \nonumber \\
                    &\quad +\left(\frac{2c_1+c_2}{2\sqrt{2}-2}\right)\frac{\|\boldeta\|_1}{m},
\end{align}
where $(a)$ is due to the guarantee on the eigenvalues of $\bSigma$ by~\cref{lem:transformed-rub-guarantee}.
By absorbing the $\sqrt{2}$ into the constant $c_1$, and $2c_1$ and $2\sqrt{2}-2$ into $c_2$, we conclude the proof. \hfill$\blacksquare$

\begin{remark}
The proof of~\cref{thm:recovery-guarantee} is similar to that of \citet{Geyer:2020}.
Indeed, this approach was used by \citet{Bruckstein:2008} to show singleton feasibility sets for compressive sensing of nonnegative signals.
That is, by applying an appropriate coordinate transformation, we can preserve recovery guarantees for an objective function that is (approximately) `flat' in the transformed space.
Then, the radius of the transformed feasible set is related to the original feasible set, \eg{} the trivial bijection between singleton sets.
\end{remark}

\begin{remark}
\Cref{thm:recovery-guarantee} differs from \citet{Geyer:2020} in that the sensing maps are rank-one, as opposed to being drawn from a Wishart distribution.
Because of this, the RIP-$\ell_2/\ell_2$ condition cannot be applied to get good sampling requirements, as pointed out by \citet{Li:2018}.
Indeed, it was shown by \citet[Lemma~2.1]{Cai:2015} that $\bigO(n^2)$ measurements are required for a rank-one sensing map to fulfill said RIP conditions sufficient for recovery of rank-one matrices ($r=1$).
This is in contrast to the $\bigO(nr)$ sampling requirements to fulfill the RUB condition used in our proof.
\end{remark}

\section{Experiments}

\begin{figure}[tb]
  \centering
  \resizebox{\linewidth}{!}{\begin{tikzpicture}

  \begin{groupplot}[
    group style={
        group name=myplots,
        group size=4 by 1,
        horizontal sep=2cm,
    },
    every axis label/.append style={
        font=\Large,
    },
    title style={
        font=\Large,
    },
    tick align=inside,
    tick pos=both,
    xmajorgrids,
    xticklabel style={
        /pgf/number format/fixed,
        /pgf/number format/precision=2,
    },
    ymajorgrids,
    yticklabel style={
        /pgf/number format/fixed,
        /pgf/number format/precision=2,
    },
    scaled y ticks=false,
    width=0.5\textwidth,
    height=0.5\textwidth,
    colorbar horizontal,
    ]

    \nextgroupplot[
    colormap={WB}{color=(white) color=(black)},
    colorbar,
    view={0}{90},
    grid=none,
    xlabel={Samples $m$},
    ylabel={Matrix size $n$},
    xlabel style={sloped like x axis},
    title={$\mathrm{rank}(\mathbf{X}_0)=1$},
    xtickmax=299, 
    xmin=25,
    ymin=15,
    ]
    \addplot3[surf,shader=faceted] table[x=m, y=n, z=err] {data/rankone.dat};
    \addplot3+[
    domain=0:23.9,
    samples=100,
    samples y=0,
    black, solid, semithick,
    no marks,
    ]
    ({x*(x+1)/2},{x},{1});
    \node at (axis cs:200,30) {\Large $m=\frac{n(n+1)}{2}$};

    \nextgroupplot[
    colormap={WB}{color=(white) color=(black)},
    colorbar,
    view={0}{90},
    grid=none,
    xlabel={Samples $m$},
    ylabel={Matrix size $n$},
    xlabel style={sloped like x axis},
    title={$\mathrm{rank}(\mathbf{X}_0)=3$},
    xmin=75,
    ymin=15,
    ]
    \addplot3[surf,shader=faceted] table[x=m, y=n, z=err] {data/lowrank.dat};
    \addplot3+[
    domain=0:41.9,
    samples=100,
    samples y=0,
    black, solid, semithick,
    no marks,
    ]
    ({x*(x+1)/2},{x},{1});
    \node at (axis cs:600,45) {\Large $m=\frac{n(n+1)}{2}$};
    
    \nextgroupplot[
    colormap={WB}{color=(white) color=(black)},
    point meta min=-7.3,
    point meta max=0,
    colorbar,
    view={0}{90},
    grid=none,
    xlabel={Samples $m$},
    ylabel={Noise Bound $\epsilon$},
    xlabel style={sloped like x axis},
    title={Noisy Measurements},
    xmin=150,
    xtickmin=151, 
    xtickmax=499, 
    xmode=normal,
    ymode=log,
    colorbar style={
        xticklabel={$10^{\pgfmathprintnumber{\tick}}$},
        x tick label style={
            /pgf/number format/.cd,
                fixed,
                precision=1,
            /tikz/.cd
        },
    },
    ]
    \addplot3[surf,shader=faceted] table[x=m, y=noise, z expr=log10(\thisrow{err})] {data/noisy.dat};
    
    \nextgroupplot[
    colormap={WB}{color=(white) color=(black)},
    point meta min=0,
    point meta max=1,
    colorbar,
    view={0}{90},
    grid=none,
    xlabel={Samples $m$},
    ylabel={Matrix size $n$},
    xlabel style={sloped like x axis},
    title={$\mathrm{rank}(\mathbf{X}_0)=n$},
    xtickmax=2999, 
    xmin=250, xmax=3000,
    ymin=20, ymax=400,
    colorbar/draw/.append code={%
        \pgfplotsset{
            colorbar sampled line={
                colormap/hot2,
                samples at={0.1, 0.2, 0.3, 0.4,
                            0.5, 0.6, 0.7, 0.8, 0.9},
                scatter,
                only marks,
                mark=|, 
                mark size=\pgfkeysvalueof{/pgfplots/colorbar/width}/2,
                ultra thick,
            },
            hide axis,
            colorbar/draw,
        }
    },
    ]
    \addplot3[surf,shader=faceted] table[x=m, y=n, z=err] {data/fullrank.dat};
    \addplot3[
    colormap/hot2,
    contour gnuplot={
        levels={0.1, 0.2, 0.3, 0.4,
                0.5, 0.6, 0.7, 0.8, 0.9},
        labels=false,
    },
    thick,
    ]
    table[x=m, y=n, z=err] {data/fullrank.dat};
    \addplot3+[
    domain=0:76.9,
    samples=100,
    samples y=0,
    black, solid, semithick,
    no marks,
    ]
    ({x*(x+1)/2},{x},{1});
    \node at (axis cs:2000,90) {\Large $m=\frac{n(n+1)}{2}$};
    
  \end{groupplot}
  
\end{tikzpicture}

}
  \caption{
  Recovery of low-rank matrices from rank-one measurements.
  \textbf{(Left)} Rate of successful recovery over 10 trials for a rank-one matrix, corresponding to the phase retrieval problem.
  Observe a linear boundary for perfect recovery, in accordance with \cref{cor:lowrank-recovery-guarantee}.
  \textbf{(Center-left)} Rate of successful recovery over 10 trials for a rank-three matrix.
  Again, there is a linear relationship between the matrix size and the number of samples needed for guaranteed recovery.
  Moreover, the slope of this boundary is one third that of the rank-one case, due to the requirement $m\in\bigO(nr)$.
  \textbf{(Center-right)} Error in recovering a fixed low-rank matrix under measurement noise.
  Observe that in the described setting, the recovery error is essentially independent of $m$ when $m$ is sufficiently large, due to the fundamental limit imposed by $\epsilon$.
  \textbf{(Right)} Error in recovering a full-rank matrix.
  Since the matrix $\bX_0$ is full-rank for all sizes $n$, perfect recovery is only attained when $m\in\bigO(n^2)$.
  However, due to the fast decay of the eigenvalues, the square-root of the Frobenius norm error is proportional to $n/m$ in the undersampled regime.
  Indeed, the slopes of the level sets (contour lines shown) scale according to the square-root of the error, as expected.
  }
  \label{fig:exp}
\end{figure}
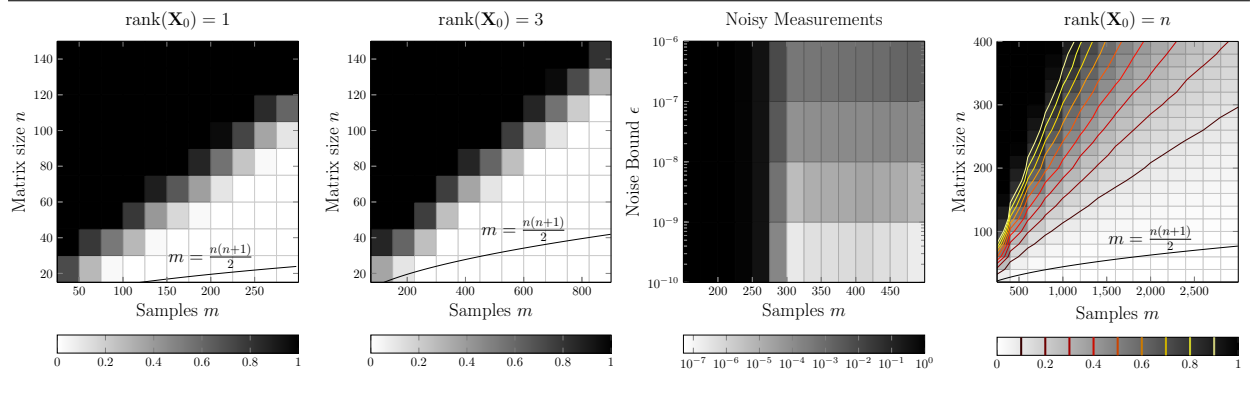

By considering matrix sensing as a simple feasibility problem, any optimization algorithm capable of finding feasible points is applicable.
In the following experiments, we apply projection methods for conic optimization~\citep{Henrion:2012}.
In particular, we consider the Lagrange dual of the affine constraint $\cZ(\bX)=\bb$, and apply the L-BFGS algorithm in the dual space to project an initial point to the intersection of the PSD cone and the affine constraint set~\citep{Bonnans:2006, Malick:2004}
\footnote{
    Implementation details can be found in \cref{app:methods}, as well as a comparison with other methods.
    Code can be found at \url{https://git.roddenberry.xyz/feasible-rop/}.
}.
This is repeated for ten trials, with independently drawn measurement vectors and random symmetric initial points for each trial.
By projecting different random matrices onto the feasible set for each trial, we avoid any hidden regularization that may occur when projecting, say, the all-zeros matrix onto the feasible set $F$.
This allows us to more accurately measure the size of the feasible set.

In each experiment, we consider $\bX_0$ to be a diagonal matrix, \ie{}, with the positive eigenvalues of $\bX_0$ on the diagonal.
This is without loss of generality: Suppose our sensing vectors are \iid{} uniformly distributed on the unit sphere.
Denote the eigendecomposition of $\bX_0$ by $\bX_0=\bU\bLambda\bU^\top$, where $\bU$ is an orthogonal matrix and $\bLambda$ is a diagonal matrix.
Then, we have
\begin{equation}
    \bz^\top\bX_0\bz=(\bU^\top\bz)^\top\bLambda(\bU^\top\bz)\overset{d}{=}\bz^\top\bLambda\bz,
\end{equation}
where $\overset{d}{=}$ indicates equality in distribution.

\subsection{Linear Number of Measurements Needed for Exact Recovery of Low-Rank Matrices}

We first consider the recovery of low-rank matrices from rank-one projections.
In~\letterref{fig:exp}{ (Left)}, we construct a rank-one matrix $\bX_0$ with unit operator norm.
Then, for each matrix size $n$, we draw a varying number of $m$ measurement vectors that are \iid{} from a multivariate normal distribution, and project a random symmetric matrix onto the feasible set $F=\{\bX\succeq 0:\cZ(\bX)=\bb\}$.
We then plot the empirical likelihood that the Frobenius norm error between the recovered matrix and the true matrix is greater than $\epsilon=10^{-3}$.

One can see a clear linear boundary between the perfect and imperfect recovery regions, in accordance with the linear sampling requirement of~\cref{cor:lowrank-recovery-guarantee} and~\cref{thm:phase-retrieval}.

To demonstrate our sampling requirements beyond the rank-one case, we repeat the same experiment for rank-three matrices, where each of the nonzero eigenvalues take unit value.
The results of this are shown in \letterref{fig:exp}{ (Center-left)}, where the same linear boundary phenomena is observed.
Additionally, the slope of this boundary is lower than the rank-one case, due to the sampling requirement of~\cref{cor:lowrank-recovery-guarantee} being linear in the rank of the matrix.

\subsection{Stability Under Measurement Noise}

To evaluate the behavior of the approximate feasible set under measurement noise, we fix a rank-three matrix $\bX_0\in\Snn{50}$ where each of the nonzero eigenvalues has unit value.
Then, we take noisy measurements $\bb=\cZ(\bX_0)+\boldeta$, where each element of $\boldeta$ is \iid{} uniformly distributed in the interval $[-\epsilon,+\epsilon]$ for some parameter $\epsilon>0$, so that $\|\boldeta\|_1\leq m\epsilon$.
In the regime where $m \gtrsim rn = 150$, then, we expect the error of recovery to only be related to the parameter $\epsilon$, and not determined by $m$~[\textit{cf.}~\eqref{eq:recovery-guarantee}].

Plotting the average Frobenius norm error in \letterref{fig:exp}{ (Center-right)}, we see that this is indeed the case.
Beyond a sufficient number of samples to capture the low-rank structure of the matrix, it is difficult for an increased number of samples to shrink the feasible set due to measurement noise.
Moreover, when $m\approx 275$, we can see that the error depends on both the number of samples and the noise level.
This indicates that there is a transition after which measurement noise dominates sampling noise, as expected by~\cref{thm:recovery-guarantee}.

\subsection{Feasible Set Shrinks Quickly When the Spectrum Decays Quickly}

We now demonstrate how simply choosing a feasible point in PSD matrix sensing is effective even when the matrix being recovered is only effectively (but not strictly) low-rank.
That is, if the eigenvalues of the matrix decay fast enough, even if all of them are nonzero, a good estimate can still be obtained without regularization.
In particular, we consider a strictly positive definite matrix $\bX_0\in\Snn{n}$ whose eigenvalues $\{\lambda_i\}_{i=1}^n$ are
\begin{equation}\label{eq:exp-fullrank-eigvals}
    \lambda_i=\frac{1}{i^{3/2}}-\frac{1}{(i+1)^{3/2}}.
\end{equation}
Notably, as $n$ grows large, the nuclear norm of $\bX_0$ quickly approaches $1$ from the left.
Moreover, the nuclear norm after removing the $r$ leading eigenvalues is approximately
\begin{equation}\label{eq:exp-fullrank-decay}
    \|[\bX_0]_{-(n-r)}\|_*\approx \frac{1}{(r+1)^{3/2}}.
\end{equation}
Applying~\cref{thm:recovery-guarantee}, then, yields a bound on the Frobenius norm error when $m\in\bigO(nr)$.
Specifically, we have a high-probability guarantee that any PSD matrix $\bX$ fitting the measurements will satisfy
\begin{equation}\label{eq:exp-fullrank-bound}
    \|\bX-\bX_0\|_F\lesssim\frac{1}{r^2}.
\end{equation}
In the context of~\cref{thm:recovery-guarantee}, $r$ is determined by the relationship between the matrix size $n$ and the number of samples $m$.
In particular, $r\lesssim m/n$, so that the level sets of the Frobenius norm error according to~\eqref{eq:exp-fullrank-bound} are expected to behave according to 
\begin{equation}\label{eq:exp-fullrank-levelsets}
   n\approx m\sqrt{\|\bX-\bX_0\|_F}.
\end{equation}

This is demonstrated in~\letterref{fig:exp}{ (Right)}, where we depict the average Frobenius norm error between the recovered matrix (again, via projection of a random matrix onto $F$) and the matrix $\bX_0$.
We plot the average error rather than the recovery rates, since there is no hope of exactly recovering the full-rank $\bX_0$ when $m\ll n^2$.
One can see that the level sets (plotted by the contour lines) of the error show a linear relationship between $m$ and $n$, as expected by~\eqref{eq:exp-fullrank-levelsets}.
Moreover, as the error increases, the slope of the level set increases according to the square-root of the error, again following~\eqref{eq:exp-fullrank-levelsets}.

\section{Conclusion}

In this work, we characterize the feasible set of the PSD matrix sensing problem with rank-one sensing maps.
In particular, we show that under sufficient sampling conditions that endow the sensing map with a suitable RUB property, the radius of the feasible set can be bounded in terms of the decay in the spectrum of the matrix and the $\ell_1$-norm of the measurement noise.
An immediate consequence of this is an $\bigO(nr)$ sampling rate for solution uniqueness when sensing PSD matrices in the absence of regularization, which matches related work considering explicit or implicit regularization techniques.
We then discuss several applications of our results, before demonstrating them on low-rank and approximately low-rank matrices.

\section*{Acknowledgements}

SS acknowledges funding by the NSF (\href{https://nsf.gov/awardsearch/showAward?AWD_ID=2008555}{CCF-2008555}).
AK acknowledges funding by the NSF (\href{https://www.nsf.gov/awardsearch/showAward?AWD_ID=1907936}{CCF-1907936}).
AK thanks TOOL for the song ``Fear Inoculum.’'
This work was partially done as TMR’s class project for ``COMP545: Advanced Topics in Optimization,'' Rice University, Spring 2020.

\newpage

\begin{appendices}

\crefalias{section}{appendix}

\section*{Appendix}

In this appendix, we establish intermediate results used in the proof of \cref{thm:recovery-guarantee}.
We also discuss the Restricted Uniform Boundedness (RUB) property~\citep{Cai:2015}, as well as its relationship with symmetric rank-one projections.
\Cref{app:srop,app:srub} are not intended to prove any new results: we gather the work by~\citet{Cai:2015} in a convenient way, with the primary goal of establishing \cref{lem:transformed-rub-guarantee}.
We refer the reader to said paper for an in-depth discussion on rank-one matrix projections and the RUB property.
After the proofs, we describe several algorithms for finding feasible points.
Although our main result provides theoretical guarantees on the solutions found by any algorithm, we compare the different approaches based on how quickly they converge to a feasible point, \ie{} a PSD matrix satisfying a set of affine measurements.

\section{Proof of \Cref{lem:fixed-trace}}
\label{app:flat-nuc}

Let an arbitrary $\bY\in\widehat{F}$ be given.
Then, there exists $\bX\in F$ such that $\bY=g(\bX)=\bV^\top\bX\bV$.
The trace of $\bY$ can then be calculated directly:
\begin{equation}\label{eq:first-trace-calculation}
  \begin{aligned}
    \trace(\bY) &= \trace(\bV^\top\bX\bV) = \trace(\bV\bV^\top\bX) \\
                &= \trace(\frac{1}{m}\sum_{i=1}^m\bz_i\bz_i^\top\bX) = \frac{1}{m}\sum_{i=1}^m\trace(\bz_i\bz_i^\top\bX) \\
                &= \frac{1}{m}\sum_{i=1}^m\bz_i^\top\bX\bz_i = \frac{1}{m}\sum_{i=1}^m\cZ_i(\bX).
  \end{aligned}
\end{equation}

Since $\bX\in F$, we have that $\sum_{i=1}^m|\cZ_i(\bX)-b_i|\leq\|\boldeta\|_1$.
This implies that
\begin{equation}
    -\|\boldeta\|_1\leq\sum_{i=1}^m\cZ_i(\bX)-\sum_{i=1}^m b_i\leq\|\boldeta\|_1.
\end{equation}
Substituting this into \eqref{eq:first-trace-calculation} yields
\begin{equation}
    \frac{\sum_{i=1}^m b_i-\|\boldeta\|_1}{m}
    \leq\trace(\bY)
    \leq\frac{\sum_{i=1}^m b_i+\|\boldeta\|_1}{m}.
\end{equation}

Moreover, since $\bY\succeq 0$ by definition of $\widehat{F}$, the trace is equal to the nuclear norm, concluding the proof.

\section{Symmetric Rank-One Projections}
\label{app:srop}

Let $\cZ:\Snn{n}\to\Rn{m}$ be a linear map of the form
\begin{equation}\label{eq:symmetric-sensing-map}
    \cZ_i(\bX)=\bz_i^\top\bX\bz_i,
\end{equation}
for $\{\bz_i\}_{i=1}^m$ independently drawn from a spherically symmetric distribution, \eg{} a multivariate normal distribution.
That is, each element of $\cZ(\bX)$ is a symmetric quadratic form on $\bX$.
For a sensing map $\cZ$ of the form~\eqref{eq:symmetric-sensing-map}, there is a natural induced sensing map.
Without loss of generality (in the noiseless case), assume $\bz$ are \iid{} samples from a multivariate normal distribution.
Define $\cZ':\Snn{n}\to\Rn{\lfloor m/2 \rfloor}$ as
\begin{equation}\label{eq:asymmetric-sensing-map}
    \cZ'_i(\bX)=
    \frac{1}{2}\left(\bz_{2i-1}+\bz_{2i}\right)^\top\bX\left(\bz_{2i-1}-\bz_{2i}\right)=
    \frac{1}{2}\left(\cZ_{2i-1}(\bX)-\cZ_{2i}(\bX)\right).
\end{equation}
As noted in the proof of \citet[Proposition~2.3]{Cai:2015}, $\bz_{2i-1}+\bz_{2i}$ and $\bz_{2i-1}-\bz_{2i}$ are \iid{} random vectors, so that $\cZ'$ follows the rank-one projection (ROP) model of \citet{Cai:2015}.

Using the measurements of $\cZ'$ is strictly less informative than using those of $\cZ$.
That is, for a symmetric matrix $\bX_0\in\Snn{n}$, and any matrix $\bX\in\Snn{n}$
\begin{equation}\label{eq:measurement-restrictiveness}
    \cZ(\bX)=\cZ(\bX_0)\rightarrow \cZ'(\bX)=\cZ'(\bX_0).
\end{equation}
This yields recovery guarantees for nuclear norm minimization subject to measurements taken via $\cZ$, stemming from the RUB properties of $\cZ'$ under sufficient sampling conditions, as in \citet[Proposition~2.3]{Cai:2015}.

\section{Symmetrized Restricted Uniform Boundedness}
\label{app:srub}

Recovery guarantees subject to rank-one measurement constraints are attained via the RUB property in \citet{Cai:2015}, defined as follows:
\begin{defn}[{\citet[Definition~2.1]{Cai:2015}}]\label{defn:rub}
  A linear sensing map $\cA:\Rnn{n}\to\Rn{m}$ is said to satisfy the \emph{Restricted Uniform Boundedness property} of order $r$ with constants $C_1,C_2$ if for all nonzero rank-$r$ matrices $\bX\in\Rnn{n}$,
  \begin{equation}\label{eq:rub-definition}
    C_1\leq\frac{\|\cA(\bX)\|_1/m}{\|\bX\|_F}\leq C_2.
  \end{equation}
  We denote the set of all such maps by $\RUB{r}{C_1}{C_2}$.
\end{defn}

For a symmetric rank-one sensing map $\cZ$, \citet{Cai:2015} use the induced asymmetric sensing map $\cZ'$, leverage their guarantees for recovery subject to the asymmetric map, then note that the original symmetric map $\cZ$ is more restrictive~\eqref{eq:measurement-restrictiveness}, and thus inherits the recovery guarantees of $\cZ'$.
For convenience, we name the set of all such \emph{symmetric} sensing maps whose induced asymmetric maps obey a sufficient RUB property:
\begin{defn}[\cref{defn:srub} redux]\label{defn:srub-redux}
    A symmetric sensing map $\cZ:\Snn{n}\to\Rn{m}$ is said to fulfill the \emph{Symmetrized Restricted Uniform Boundedness property} (SRUB) of order $r$ with constants $C_1,C_2$ if the induced asymmetric sensing map~\eqref{eq:asymmetric-sensing-map} $\cZ'$ has the RUB property of order $r$ with constants $C_1,C_2$, \ie{} $\cZ'\in\RUB{r}{C_1}{C_2}$.
    We denote the set of all such symmetric sensing maps by $\SRUB{r}{C_1}{C_2}$.
\end{defn}

\section{Proof of \Cref{prop:basic-sing-sampling}}
\label{app:sing-proof}

Observing that $\bSigma$ is equivalent to the sample covariance matrix of $m$ \iid{} random vectors distributed uniformly on the unit sphere, the population covariance matrix as $m\to\infty$ is the identity matrix.
We establish the rate at which $\bSigma\to\bI$ by leveraging the result of~\citet[Corollary~5.52]{Vershynin:2010}, which states
\begin{theorem}[{\citet[Corollary~5.52]{Vershynin:2010}}]\label{thm:sample-convergence}
    Consider a distribution in $\Rn{n}$ with covariance matrix $\bC$ and supported in some centered Euclidean ball with radius $\sqrt{r}$.
    Let $\epsilon\in(0,1)$ and $t\geq 1$.
    Then, if
    \begin{equation}
        m\geq C\left(\frac{t}{\epsilon}\right)^2\|\bC\|_2^\inv r\log n
    \end{equation}
    for some absolute constant $C$, then
    \begin{equation}
        \|\widehat{\bC}-\bC\|_2\leq\epsilon\|\bC\|_2
    \end{equation}
    with probability at least $1-n^{-t^2}$, where $\widehat{\bC}$ is the sample covariance of $m$ samples from the distribution.
\end{theorem}

In the context of \cref{thm:sample-convergence}, we have $r=1$ and $\bC=\bI$.
So, if $m\geq C(t/\epsilon)^2\log n$, then $1-\epsilon\leq\sigmin(\bSigma)\leq\sigmax(\bSigma)\leq 1+\epsilon$ with high probability.
Seeking to bound their ratio:
\begin{equation}
  \begin{aligned}
    \frac{\sigmax(\bSigma)}{\sigmin(\bSigma)} &\leq \frac{1+\epsilon}{1-\epsilon} \leq \sqrt{2} \\
    1+\epsilon &\leq (1-\epsilon)\sqrt{2} \\
    \epsilon &\leq \frac{\sqrt{2}-1}{\sqrt{2}+1}\approx 0.1716.
  \end{aligned}
\end{equation}
So, if $\epsilon\leq 0.1716$, the ratio of the extreme singular values of $\bSigma$ will be bounded by $\sqrt{2}$, with high probability.
Choosing an appropriate value of $\epsilon$, along with $t=\sqrt{n\gamma/\log n}$, yields this condition with probability at least $1-\exp(-n\gamma)$ when $m\in\bigO(n)$, as desired.

\section{Proof of \Cref{lem:transformed-rub-guarantee}}
\label{app:trans-srub}

Let $\cZ:\Snn{n}\to\Rn{m}$ be a symmetric, rank-one sensing map with sensing vectors $\{\bz_i\}_{i=1}^m$ drawn \iid{} uniformly on the unit sphere.
Suppose that $\cZ\in\SRUB{2kr}{C_1}{C_2}$ with $C_1,C_2,k>0$ such that $C_2/C_1<\sqrt{k}$.
Therefore, the induced asymmetric sensing map satisfies $\cZ'\in\RUB{2kr}{C_1}{C_2}$.
That is, for any rank-$2kr$ matrix $\bX\in\Snn{n}$,
\begin{equation}\label{eq:rub-example}
    C_1\leq\frac{\|\cZ'(\bX)\|_1/\lfloor m/2\rfloor}{\|\bX\|_F}\leq C_2.
\end{equation}

For an invertible matrix $\bV\in\Rnn{n}$, define $\widehat{\cZ}:\Snn{n}\to\Rn{m}$ as before, \ie{}
\begin{equation}\label{eq:transformed-coordinate}
    \widehat{\cZ}(\bY)=(\bV^\inv\bz_i)^\top\bY(\bV^\inv\bz_i).
\end{equation}
Observe that this transformed sensing map simply applies $\bV^\inv$ to each sensing vector.
In particular, for matrices $\bY\in\Snn{n}$, and $\bX\in\Snn{n}$ such that $\bY=\bV^\top\bX\bV$ and $\rank(\bX)=\rank(\bY)$, we have for all $i=1,\ldots,m$
\begin{equation}\label{eq:transformed-equivalence}
    \widehat{\cZ}_i(\bY) = \widehat{\cZ}_i(\bV^\top\bX\bV) = \bz_i^\top\bX\bz_i = \cZ_i(\bX).
\end{equation}  
That is, for any matrix $\bY\in\Snn{n}$, $\widehat{\cZ}(\bY)$ can be written as $\cZ(\bX)$ for some related matrix $\bX\in\Snn{n}$ of the same rank as $\bY$.
Therefore, the induced asymmetric sensing map enjoys the same equivalence:
\begin{equation}\label{eq:transformed-induced-equivalence}
    \widehat{\cZ'}(\bY) = \cZ'(\bX).
\end{equation}

We now establish the RUB properties of $\widehat{\cZ'}$ relative to those of $\cZ'$.
First, for any rank-$2kr$ matrix $\bY\in\Snn{n}$, there exists a rank-$2kr$ matrix $\bX\in\Snn{n}$ such that $\bY=\bV^\top\bX\bV$, thus
\begin{equation}\label{eq:transformed-rub-relationship}
    \frac{\|\widehat{\cZ'}(\bY)\|_1/\lfloor m/2\rfloor}{\|\bY\|_F}=
    \frac{\|\cZ'(\bX)\|_1/\lfloor m/2\rfloor}{\|\bV^\top\bX\bV\|_F}.
\end{equation}
Since we assume that $\cZ'\in\RUB{2kr}{C_1}{C_2}$, by~\cref{defn:rub} we get
\begin{equation}\label{eq:transformed-rub}
    \frac{C_1}{\sigmax^2(\bV)}\leq
    \frac{\|\cZ'(\bX)\|_1/\lfloor m/2\rfloor}{\|\bV^\top\bX\bV\|_F}\leq
    \frac{C_2}{\sigmin^2(\bV)}.
\end{equation}
Substituting~\eqref{eq:transformed-rub-relationship} into~\eqref{eq:transformed-rub} and applying~\cref{defn:srub} yields
\begin{equation}
    \widehat{\cZ}\in\SRUB{2kr}{C_1/\sigmax^2(\bV)}{C_2/\sigmin^2(\bV)}.
\end{equation}

The assumed SRUB property of $\cZ$ is attained for $m\geq Cnr$ for some sufficiently large constant $C$ by \cref{prop:basic-rub-sampling}, with high probability, since we can choose $C_1,C_2,k$ so that $C_2/C_1<\sqrt{k}$.
Then, to ensure
\begin{equation}
    \frac{C_2/\sigmin^2(\bV)}{C_1/\sigmax^2(\bV)}=\frac{C_2\sigmax(\bSigma)}{C_1\sigmin(\bSigma)}<\sqrt{2k},
\end{equation}
the extreme singular values of $\bSigma$ must have a ratio less than $\sqrt{2}$.
\Cref{prop:basic-sing-sampling} guarantees this when $m\geq Cn$ for some sufficiently large constant $C$, with high probability.
Under these conditions, $\widehat{\cZ}$ satisfies the desired SRUB property with probability at least $1-\bigO(\exp(-n\gamma))$, as desired.

\section{Proof of \eqref{eq:lower-nuclear-bound}}
\label{app:lower-nuc}

For a matrix $\bC\in{\mathbb C}^{n\times m}$, let the singular values of $\bC$ be denoted by $\sigma_1(\bC)\geq\ldots\geq\sigma_n(\bC)$.
We leverage a result of \citet{Wang:1997}:
\begin{theorem}[\citet{Wang:1997}]\label{thm:singular-product}
    Let $\bA\in{\mathbb C}^{n\times n}$, $\bB\in{\mathbb C}^{n\times m}$, and $1\leq i_1<\ldots<i_k\leq n$, $0<p\in\R$.
    Then
    \begin{equation}\label{eq:singular-product}
        \sum_{t=1}^k\sigma_{i_t}^p(\bA\bB)\geq\sum_{t=1}^k\sigma_{i_t}^p(\bA)\sigma_{n-t+1}^p(\bB). 
    \end{equation}
\end{theorem}

In particular, let $i_t=r+t$ for $t=1,\ldots,n-r$, and let $p=1$.
Then, since $\bX_0=\bV^\invt\bY_0\bV^\inv$,
\begin{equation}
  \begin{aligned}
    \sum_{t=1}^{n-r}\sigma_{i_t}(\bX_0) &= \sum_{t=1}^{n-r} \sigma_{i_t}(\bV^\invt\bY_0\bV^\inv) \\
    &\overset{(a)}{\geq}  \sum_{t=1}^{n-r} \sigma_{i_t}(\bV^\invt\bY_0) \sigma_{n-t+1}(\bV^\inv) \, \geq \, \sigmin(\bV^\inv) \sum_{t=1}^{n-r} \sigma_{i_t}(\bV^\invt\bY_0) \\
    &\overset{(b)}{\geq} \sigmin(\bV^\invt) \sum_{t=1}^{n-r} \sigma_{i_t}(\bY_0) \sigma_{n-t+1}(\bV^\invt) \, \geq \, \sigmin^2(\bV^\invt) \sum_{t=1}^{n-r} \sigma_{i_t}(\bY_0) \\
    &\overset{(c)}{=} \sigmax^\inv(\bSigma) \sum_{i=r+1}^n \sigma_i(\bY_0),
  \end{aligned}
\end{equation}
where $(a),(b)$ are due to~\cref{thm:singular-product}, and $(c)$ is a consequence of the factorization $\bSigma=\bV\bV^\top$.
Therefore,
\begin{equation}
    \|[\bY_0]_{-(n-r)}\|_*\leq\sigmax(\bSigma)\|[\bX_0]_{-(n-r)}\|_*,
\end{equation}
as desired.

\section{Semidefinite Projection Methods}
\label{app:methods}

In this section, we consider a few methods for finding feasible points for the matrix sensing problem, then compare their performance on a small example.
We begin by briefly describing the approach of each algorithm: for a full discussion, see the referenced papers.

\subsection{Dual Space Methods}

\citet{Henrion:2012} propose projection methods coupled with a dual space optimization procedure for finding feasible points in conic optimization.
In particular, for some matrix $\bC\in\Snn{n}$, let the program we solve be
\begin{argmini}
    {\bX\succeq 0}{\frac{1}{2}\|\bC-\bX\|_F^2}
    {\label{eq:proj-program}}{\bX^* \in }
    \addConstraint{\cZ(\bX)}{=\bb.}
\end{argmini}
By taking the Lagrangian with respect to the affine constraints, we get the dual function
\begin{mini}
    {\bX\succeq 0}{\frac{1}{2}\|\bC-\bX\|_F^2-\by^\top(\cZ(\bX)-\bb).}
    {\label{eq:proj-dual}}{\theta(\by)=}
\end{mini}
They show that minimizing $\bX$ has a closed-form expression, yielding $\theta(\by)$ and its gradient as closed-form expressions as well:
\begin{equation}\label{eq:proj-dual-closed}
    \begin{aligned}
        \bX(\by) &= \cP_{psd}(\bC+\cZ^\top(\by)) \\
        \theta(\by) &= \by^\top\bb + \frac{1}{2}(\|\bC\|_F^2-\|\bX(\by)\|_F^2) \\
        \nabla\theta(\by) &= -\cZ(\bX(\by)) + \bb.
    \end{aligned}
\end{equation}

Maximizing the concave function $\theta(\by)$, then, can be done using any gradient-based method.
In particular, the authors suggest the use of quasi-Newton methods, such as L-BFGS with Wolfe line search~\citep{Bonnans:2006}.

\subsection{Nesterov's Method}

As done by \citet{Demanet:2014}, we use the distance from the measurements as a loss function:
\begin{equation}\label{eq:nest-criterion}
    f(\bX)=\frac{1}{2}\|\cZ(\bX)-\bb\|_2^2.
\end{equation}
And use Nesterov iterations with a positive semidefinite projection and stepsize $\eta$:
\begin{align}
    \bX_0 &= \bY_0 = 0 \\
    \bX_k &= \cP_{psd}(\bY_{k-1}-\eta\nabla_g(\bY_{k-1})) \\
    \theta_k &=  2\left(1 + \sqrt{1 + 4/\theta_{k-1}^2}\right)^\inv \\
    \beta_k &=  \theta_k(\theta_{k-1}^\inv -1) \\
    \bY_k &= \bX_k + \beta_k(\bX_k - \bX_{k-1}).
\end{align}

\subsection{Douglas-Rachford Splitting}

As suggested by \citet{Demanet:2014}, we write the feasibility condition as a split loss function:
\begin{equation}\label{eq:dr-criterion}
    f(\bX)=\iota_{\cZ(\bX)=\bb}(\bX)+\iota_{\bX\succeq 0}(\bX),
\end{equation}
where $\iota$ is an indicator taking value $0$ within a set and $\infty$ outside of a set.
Then, we can simply apply the Douglas-Rachford algorithm~\citep{Douglas:1956}:
\begin{align}
    \bX_0 &= \bY_0 = 0 \\
    \bY_k &= \cP_{\cZ(\bX)=\bb}(2\bX_{k-1}-\bY_{k-1})-\bX_{k-1}+\bY_{k-1} \\
    \bX_k &= \cP_{psd}(\bY_k).
\end{align}

\subsection{Factored Gradient Descent (FGD)}

When the rank of the solution is known, or at least bounded, factorized methods can enforce low-rankness and positive semidefiniteness by construction.
In particular, we recover the matrix $\bX_0$ by optimizing over a `tall' matrix factor $\bU$:
\begin{argmini}
    {\bU\in\Rnp{n}{r}}{\frac{1}{2}\|\cZ(\bU\bU^\top)-\bb\|_2^2.}
    {\label{eq:fact-program}}{\bU^* \in }
\end{argmini}

Following the iteration scheme of \citet{Park:2016},
\begin{align}
    \bU_0\bU_0^\top &\approx \cZ^*(\bb) \\
    \bX_k &= \bU_k\bU_k^\top \\
    \bU_k &= \left(\bI - \eta\nabla f(\bX_{k-1})\right)\bU_{k-1},
\end{align}
for some stepsize $\eta>0$, and using the least-squares loss $f(\bX)=\frac{1}{2}\|\cZ(\bX)-\bb\|_2^2$, we recover the matrix $\bX^*=\bU^*[\bU^*]^\top$.
The initialization $\bU_0\bU_0^\top$ is the best rank-$r$ PSD approximation of $\cZ^*(\bb)$, with appropriate normalization.
This initialization is necessary as opposed to setting $\bU_0=0$, which is a clear saddle point.

\subsection{Numerical Experiments}

\begin{figure}[tb]
  \centering
  \resizebox{0.8\linewidth}{!}{\begin{tikzpicture}

  \begin{axis}[
    tick align=inside,
    tick pos=both,
    xmajorgrids,
    xticklabel style={
      /pgf/number format/fixed,
      /pgf/number format/precision=2,
    },
    ymajorgrids,
    yticklabel style={
      /pgf/number format/fixed,
      /pgf/number format/precision=2,
    },
    yminorticks=true,
    scaled y ticks=false,
    width=\textwidth,
    height=0.4\textwidth,
    xlabel={Iteration},
    ylabel={$\|\cZ(\bX)-\bb\|$},
    xmode=log,
    ymode=log,
    ymin=1e-8, ymax=1e2,
    ]

    \addplot+[] table[x=iter, y=gradient] {data/lbfgs.dat};
    \addlegendentry{L-BFGS}
    \addplot+[] table[x=iter, y=loss] {data/nesterov.dat};
    \addlegendentry{Nesterov}
    \addplot+[] table[x=iter, y=loss] {data/dougrach.dat};
    \addlegendentry{Douglas-Rachford}
    \addplot+[] table[x=iter, y=loss] {data/fgdlr.dat};
    \addlegendentry{FGD ($r=1$)}
    \addplot+[] table[x=iter, y=loss] {data/fgdfr.dat};
    \addlegendentry{FGD ($r=n$)}

  \end{axis}
  
\end{tikzpicture}}
  \caption{
  Convergence of feasible point finding algorithms per iteration.
  Nesterov's method and the Douglas-Rachford approach both compute a single eigendecomposition at every iteration, and L-BFGS may compute multiple eigendecompositions due to the line search.
  FGD has no need to compute eigendecompositions, since the representation of the solution ensures it remains in the PSD cone.
  Clearly, L-BFGS converges with the fewest number of iterations, due to its use of line search techniques and second-order information.
  This is closely followed by FGD when $r=1$, due to the extra regularization imposed by enforcing low-rank solutions.
  }
  \label{fig:convergence}
\end{figure}

\begin{table}[tb]
\caption{
Runtime for each algorithm to achieve feasibility error $\|\cZ(\bX_k)-\bb\|<10^{-5}$.
Best performers in each category are \textbf{bolded}.
Although L-BFGS converged in the fewest number of iterations, the critically parameterized ($r=1$) FGD scheme had faster runtime by an order of magnitude.
This is due to the fact that this approach does not need to project matrices onto the PSD cone at any point, saving significant computational cost at each iteration.
$^*$FGD with $r=n$ did not converge to feasibility error less than $10^{-5}$ before terminating at $10000$ iterations.
}
\label{tab:runtime}
\vskip 0.15in
\begin{center}
\begin{small}
\begin{sc}
\begin{tabular}{lrrr}
\toprule
Algorithm        & Iterations  & Time (ms)    & Time (ms)/Iterations \\
\midrule
FGD ($r=1$)      & 62          & \textbf{0.456} & \textbf{0.00735} \\
FGD ($r=n$)$^*$  & 10000       & 394          & 0.0394 \\
L-BFGS           & \textbf{21} & 4.32         & 0.206 \\
Nesterov         & 440         & 73.9         & 0.168 \\
Douglas-Rachford & 2600        & 366          & 0.141 \\
\bottomrule
\end{tabular}
\end{sc}
\end{small}
\end{center}
\vskip -0.1in
\end{table}

We compare the performance of the described methods for finding a feasible point $\bX$, \ie{} $\cZ(\bX)=\bb, \bX\succeq 0$.
We generate a rank-one PSD matrix $\bX_0\in\Snn{15}$, and take $m=100$ measurements.
For all algorithms apart from FGD, we take $\bX=0$ as an initial point, and run until either the feasibility error $\|\cZ(\bX)-\bb\|_2<10^{-5}$, or until $10000$ iterations are performed%
\footnote{Code can be found at \url{https://git.roddenberry.xyz/feasible-rop/}}.

The convergence in terms of the number of iterations is plotted in~\cref{fig:convergence}.
Clearly, L-BFGS converges in the fewest number of iterations, followed by FGD when $r=1$.
However, as listed in~\cref{tab:runtime}, the critically parameterized FGD approach had better clocktime performance by an order of magnitude, since it does not need to perform eigendecompositions to project matrices onto the PSD cone.
To get such fast convergence requires express knowledge of the rank of the underlying matrix, though, so we prefer the L-BFGS approach for dual space projection in general.

The stepsizes for gradient-based methods without line search (Nesterov's method, FGD) were hand-tuned to achieve fastest convergence.
In particular, Nesterov's method used a stepsize of $\eta=0.1$, FGD with $r=1$ used a stepsize of $\eta=1.2$, and FGD with $r=n$ used a stepsize of $\eta=0.5$.

\end{appendices}

\newpage

\bibliography{ref}
\bibliographystyle{icml2020}

\end{document}